\documentclass[10pt,journal,compsoc]{IEEEtran}

\usepackage{amsmath,amsfonts}
\usepackage{amssymb}
\usepackage{algorithm}
\usepackage{array}
\usepackage[caption=false,font=normalsize,labelfont=sf,textfont=sf]{subfig}
\usepackage{textcomp}
\usepackage{stfloats}
\usepackage{url}
\usepackage{verbatim}
\usepackage{graphicx}
\usepackage{cite}
\usepackage{booktabs}       
\usepackage{nicefrac}       
\usepackage{microtype}      
\usepackage{xcolor}         
\usepackage{graphicx}
\usepackage{float} 

\usepackage{algpseudocode}

\usepackage{multirow}

\usepackage{adjustbox}

\usepackage{pifont}
\usepackage{caption}

\newcommand{\citep}[1]{{\cite{#1}}}
\newcommand{\citet}[1]{{\cite{#1}}}

\newcommand{\strategyname} {TopoNAS}
\newcommand{\todo}[1]{{\color{red} TODO: {#1}} }

\newcommand{\topone}[1]{{\color{red} \textbf{#1}} }
\newcommand{\toptwo}[1]{{\color{blue} {#1}} }

\begin{document}

\title{TopoNAS: Boosting Search Efficiency of Gradient-based NAS via Topological Simplification}

\author{Danpei Zhao, Zhuoran Liu, Bo Yuan
\thanks{This work was supported by the National Natural Science Foundation of China under Grant 62271018. Danpei Zhao, Zhuoran Liu and Bo Yuan are with the Image Processing Center, Beihang University, Beijing, 102206.}
\thanks{Manuscript received xx, 2024; revised xx, 2024.}
}

\markboth{Journal of \LaTeX\ Class Files,~Vol.~14, No.~8, August~2021}%
{Shell \MakeLowercase{\textit{et al.}}: A Sample Article Using IEEEtran.cls for IEEE Journals}

\IEEEpubid{0000--0000/00\$00.00~\copyright~2021 IEEE}

\maketitle

\begin{abstract}
Improving search efficiency serves as one of the crucial objectives of Neural Architecture Search (NAS). However, many current approaches ignore the universality of the search strategy and fail to reduce the computational redundancy during the search process, especially in one-shot NAS architectures. Besides, current NAS methods show invalid reparameterization in non-linear search space, leading to poor efficiency in common search spaces like DARTS. 
In this paper, we propose TopoNAS, a model-agnostic approach for gradient-based one-shot NAS that significantly reduces searching time and memory usage by topological simplification of searchable paths. Firstly, we model the non-linearity in search spaces to reveal the parameterization difficulties. To improve the search efficiency, we present a topological simplification method and iteratively apply module-sharing strategies to simplify the topological structure of searchable paths. In addition, a kernel normalization technique is also proposed to preserve the search accuracy.
Experimental results on the NASBench201 benchmark with various search spaces demonstrate the effectiveness of our method. It proves the proposed TopoNAS enhances the performance of various architectures in terms of search efficiency while maintaining a high level of accuracy. The project page is available at~\url{https://xdedss.github.io/topo_simplification}.

\end{abstract}


\section{Introduction}

Neural Architecture Search (NAS) improves the ability of task-driven model self-optimization, breaking through the shackles of the manually designed network. In recent years, NAS methods have focused on automated procedures to search for appropriate network structures, which often outperform manually designed structures in both accuracy and speed on multiple tasks~\cite{white2023nas1000}, and can be optimized for specific hardware~\cite{wu2019fbnet,shaw2019squeezenas,dong2018dpp}. The high effectiveness and broad adaptability of NAS has led to a wide range of applications, including classification~\cite{sun2019evolvingcnn}, semantic segmentation~\cite{autodeeplab,zhang2019customizable,huang2022adwu_medicalseg,weng2019unet}, object detection~\cite{xu2019autofpn,wang2022fastdartsdet,liang2021opanas,li2021asfd,yao2020smnas} and disparity estimation~\cite{autodispnet}, etc. 

NAS can be considered a special form of hyperparameter optimization (HPO)~\cite{yang2020hyperparameter}, to achieve the most efficient task-driven network architecture. The heuristic studies of NAS can be traced as early as the 1980s by~\citet{miller1989designing}. Practical NAS for modern neural networks start with reinforcement learning (RL) based methods~\cite{nasrl, nasnet}, which utilize RL to train a network to generate better neural architectures. However, the excessive time consumption of RL and model evaluation procedures limit the potential advancement of these methods, prompting the exploration of more efficient search strategies \cite{xie2023efficient}. Researchers have studied the performance of NAS methods through various ways including random search~\cite{li2020randomsearch}, evolutionary algorithms~\cite{real2017large,real2019regularizedrevolution,suganuma2020evolution} and parameter sharing~\cite{enas}, etc.

\begin{figure*}[htb]
\centering
\includegraphics[width=1.0\linewidth]{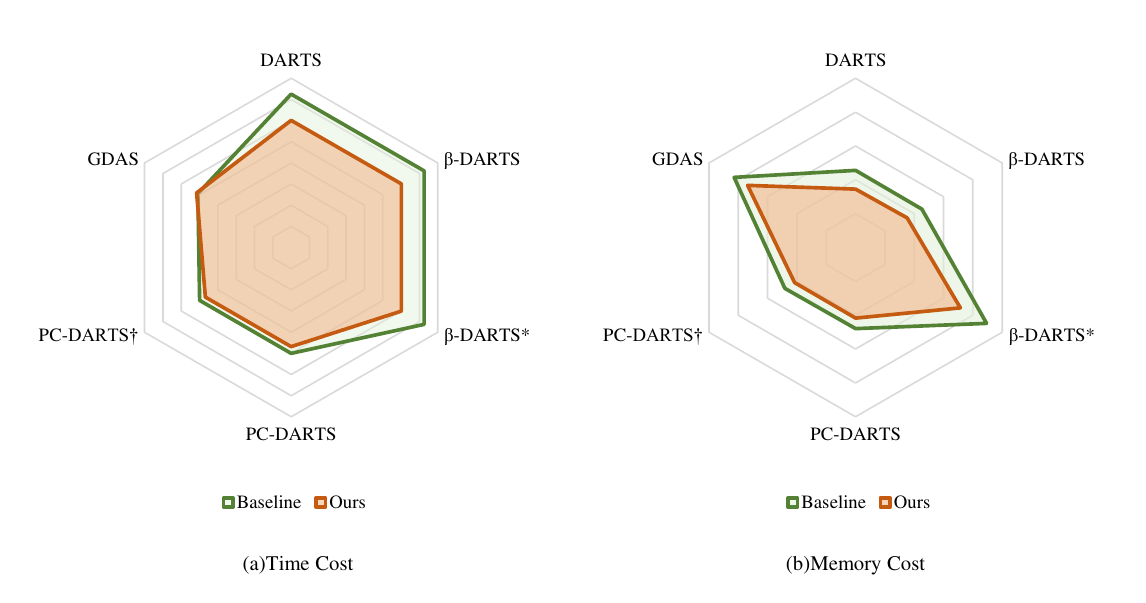}
\caption{ \strategyname{} improves search efficiency in terms of time cost and memory cost with various gradient-based NAS baselines. $\beta$-DARTS denotes searching with standard NASBench201 settings. $\beta$-DARTS* denotes searching with the settings aligned with the original work~\cite{ye2022beta}. $\dagger$ denotes combining $\beta$ regularization proposed by $\beta$-DARTS with PC-DARTS.  }
\label{fig_radar}
\end{figure*}

Recent one-shot NAS architectures including DARTS~\cite{liu2018darts} stand as a practical way to conduct efficient searching strategies. However, considering search spaces with larger supernet or more operations, the drastic increase in the cost of computational resources and time limit such forms application. Many improvements are proposed to tackle this issue. Concretely, sampling-based strategies~\cite{dong2019gdas,xu2019pc,cai2018proxylessnas} reduce the search cost by only computing a sampled part of the supernet for each iteration. Reparameterization-based methods~\cite{stamoulis2020single,wang2021mergenas,wang2022eautodet,wang2023diymerge} merge certain operations into one to simplify the supernet, reducing both search time and memory cost. However, these methods overlook the redundant computing of similar but expensive operations during the search process. There is a lack of research on the underlying topological structure of searchable paths, which can be further optimized for search efficiency.

\begin{figure*}[htb]
\centering
\includegraphics[width=0.8\linewidth]{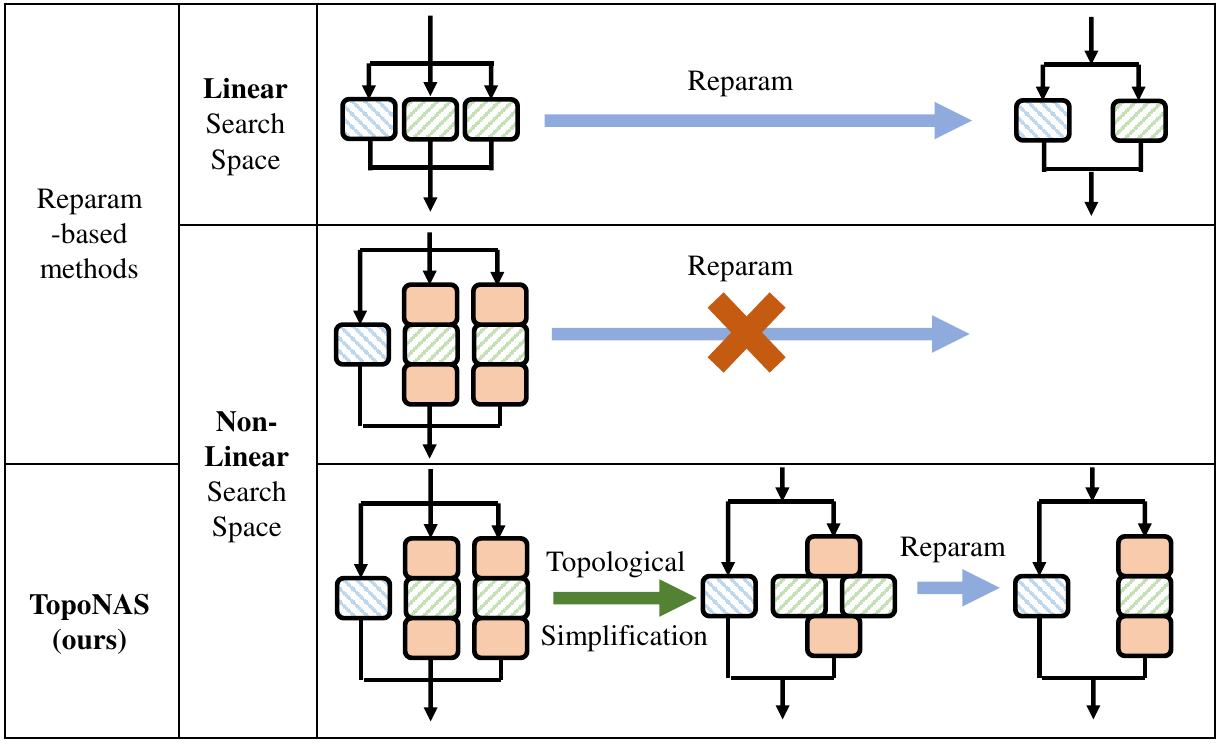}
\caption{In this graph, each rectangle represents a module to be computed during the search phase while black arrows denote the computing order. \strategyname{} enables efficient reparameterization for non-linear search spaces, where existing methods become infeasible due to unaligned levels of non-linearity. }
\label{fig_infeasible}
\end{figure*}

Through exploring the inner structure of searchable paths, we propose that the topological structure inside a searchable path includes repeated computation of similar modules that can be simplified, which provides a new aspect for accelerating the searching process. Therefore, we present a topological simplification strategy that significantly reduces the overall search cost. Firstly, we identify a critical issue of reparameterization-based NAS, which causes the degradation of the optimization process and harms the search stability. To address this issue, a kernel normalization strategy is proposed to enhance search accuracy. On top of that, we propose TopoNAS, a simple, robust, and model-agnostic searching strategy to simplify the topology of the search space and further improve search efficiency and stability. As shown in Fig. \ref{fig_radar}, \strategyname{} cooperates with various gradient-based NAS methods and achieves better search efficiency. In summary, TopoNAS possesses the following advantages: 1) Fewer Constraints: As illustrated in Fig.~\ref{fig_infeasible}, TopoNAS does not require the candidate operations to be linear or restrict the form of candidate operations. Therefore, it can be applied to any cell-based search space without special handling. 2) Better Stability: By theoretically identifying the essential reason for the search instability when reparameterization is applied along with searchable linear operations, TopoNAS comes with a kernel normalization technique that ensures a more stable search process. 3) Model-agnostic Plug-in: TopoNAS can serve as an individual plug-in that can be easily used for most state-of-the-art NAS methods to increase their efficiency. 
Experimental results on multiple datasets validate the effectiveness of the proposed model.  Experimental results also show the proposed method can bring over 9.1\% IN-16 accuracy improvement to DARTS series architectures while reducing normalized cost.

The main contribution of this paper is summarized as follows. 
\begin{itemize}
    \item We propose TopoNAS, a general searching acceleration strategy for gradient-based one-shot NAS, which acts as a plug-in that speeds up the searching process.
    \item We propose partial module sharing (PMS) and floating module sharing (FMS) to simplify the computing of the supernet, along with a customized kernel normalization approach to enhance search stability. 
    \item Experimental results on various datasets show the proposed TopoNAS can boost gradient-based one-shot NAS architectures, especially achieving the compatibility between accuracy and searching cost.
\end{itemize}

The rest of this paper is organized as follows. In Sec.\ref{s_related_works}, we review the current state of gradient-based efficient NAS strategies and analyze the strengths and limitations of recent works. Sec.\ref{s_methods} provides a thorough explanation of the proposed simplification process. In Sec.\ref{s_exp}, we carry out experiments using multiple datasets and search spaces for comparison.

\section{Related Works}  \label{s_related_works}

There are two different objectives in optimizing the search phase of NAS. One is to improve the robustness of the search and produce structures with better stability. The other is to enhance the efficiency of the search process so that the architecture can be obtained with less time and computing resource cost. Our research belongs to the latter as we are dedicated to improving search efficiency while not harming search stability and accuracy.

In this section, we first review the formulation of gradient-based NAS. Then, a brief introduction to methods aiming to improve search stability is provided. Subsequently, we introduce related researches that aim to improve NAS search efficiency, which can be grouped into sampling-based and reparameterization-based techniques. The proposed TopoNAS aims to boost the search efficiency of gradient-based one-shot NAS methods via modeling the topology of searchable paths. TopoNAS can cooperate with both sampling-based methods and reparameterization-based methods to further optimize the search process. TopoNAS is also compatible with other NAS methods focusing on improving search robustness.

\subsection{Gradient-based NAS}

NAS algorithms search for the optimal model structure in a given search space and substantially simplify the design process of deep neural networks. Early works~\cite{nasrl,nasnet} use a reinforcement learning-based framework to search for optimal network structures. These methods require a complete evaluation of each subnetwork and are therefore inefficient. To address this problem, subsequent studies enhance the searching performance by applying performance prediction~\cite{baker2017accelerating,pnas,Dong_2019_setn} and parameter sharing~\cite{enas} strategies. However, as the search space is discrete, it is hard to solve the optimization problem efficiently.

DARTS~\cite{liu2018darts} rebuilds the searching framework by introducing continuous relaxation to the search space and has been widely used in various applications~\cite{weng2019unet,peng2019efficient} for its superior efficiency. The method relaxes the cell-based discrete search space into a continuous search space and searches for the optimal solution using gradient descent, avoiding the need for repeated evaluations. Specifically, DARTS is formulated as a bi-level optimization problem:

\begin{equation}
\label{e_darts}
	\begin{split}
\min_{\alpha} & \quad  {\mathcal{L}_{val} ( w^*(\alpha), \alpha ) }
\\
\text{s.t.}  & \quad w^*( \alpha ) = \text{argmin}_{w}  \mathcal{L}_{train} ( w, \alpha )
	\end{split}
\end{equation}

\noindent where $w$ represents the weights of a supernet consisting of all possible operations and $\alpha$ represents a set of structural parameters that describes the continuous relaxed search space. $\mathcal{L}_{val}$ denotes the loss function on the validation set while $\mathcal{L}_{train}$ denotes the loss function on the training set. $ w^*( \alpha )$ denotes the optimal weights $w$ trained on the training set given a specific configuration of structural parameters. During the search, model weights and structural parameters are updated in an interleaved fashion to efficiently find the optimal structure.

\subsection{Stability Improvements}

More recent methods have improved the search accuracy by introducing modifications to the original DARTS method~\cite{chu2021fairnas,movahedi2022lambda,chu2020dartsminus,yang2021towards,ye2022beta,ye2023betapp}. Out of these methods, FairNAS~\cite{chu2021fairnas} focuses on improving fairness between searched operations by introducing strict fairness constraints. $\beta$-DARTS~\cite{ye2022beta} employs a simple yet effective $\beta$-regularization term on architecture parameters to constrain the optimization objective and exhibits strong stability. $\beta$-DARTS++~\cite{ye2023betapp} extends this with bi-level regularization to improve robustness against various proxies. $\Lambda$-DARTS~\cite{movahedi2022lambda} tackles the performance collapse problem in DARTS by harmonizing operation selection across cells, ensuring more consistent and reliable architecture discovery. However, these methods do not focus on optimizing the computing resources required for the search. DARTS-derived methods face the challenge of excess search time and GPU memory usage because they require every possible candidate operation to be stored and computed every iteration, resulting in poor performance on large search spaces. 

\subsection{Sampling-based Efficient NAS}

Sampling is a powerful tool to reduce the search cost of NAS algorithms. The early NAS methods using evolutionary algorithms~\cite{Real2017LargeScaleEO,liu2017hierarchical_evolution} can be seen as sampling-based NAS as only a small number of architectures among the search space were sampled and evaluated during the mutation process. After the emergence of one-shot NAS, efforts have been made to apply the idea of sampling to differentiable searching processes. Instead of sampling the entire subnet, more fine-grained sampling strategies are used. SPOS~\cite{guo2020spos} combines architecture sampling with parameter sharing by sampling a single path of the supernet during training steps and also adopts an evolutionary algorithm after training to find the best candidate architecture.

As for gradient-based NAS methods, ProxylessNAS~\cite{cai2018proxylessnas} creates a randomly sampled binary mask to reduce the number of paths computed during the weight update step. In the architecture update step, two paths are sampled, one of which is enhanced and the other is suppressed. GDAS~\cite{dong2019gdas} introduces Gumbel-softmax distribution to sample only one single path while retaining the gradient of architecture parameters. As the sampling probability for each operation decreases, the search process becomes more efficient, but at the same time, the risk of insufficient training is increased. Therefore, more epochs are needed for the supernet to converge. PC-DARTS~\cite{xu2019pc} takes a channel-level sampling approach by selectively inputting partial channels of feature maps into searchable operations while leaving other channels unaltered. As the number of trainable weights is decreased, PC-DARTS ease the requirement of extra searching epochs.


\subsection{Reparameterization-based Efficient NAS}

Reparameterization is widely used in convolutional neural networks to simplify and accelerate the training and inference of the network. DiracNets~\cite{zagoruyko2017diracnets} replaces residual structures in deep CNNs with simple parameterized convolutions, allowing single path networks to achieve similar performance to ResNet~\cite{resnet}. RepVGG~\cite{ding2021repvgg} demonstrates that a VGG-like network can be reparameterized to represent complex residual structures. Such a network is easy to train and inference, yet powerful enough to outperform ResNet.

In recent years, research has been conducted to combine the strength of reparameterization with NAS techniques. Single-path NAS~\cite{stamoulis2020single} shows that reparameterization can be applied to one-shot NAS in a convolution-only search space. Similarly, MergeNAS~\cite{wang2021mergenas} proposes to merge linear operations in searchable paths to reduce computational costs. With the help of the kernel merging technique, EAutoDet~\cite{wang2022eautodet} searches in a more complex detection task and can explore a larger search space within a few GPU-days. RepNAS~\cite{zhang2023repnas} is also a reparameterization-related NAS method. However, it aims to improve the inference performance of searched architecture by building the search space with re-parameterizing blocks, while the search cost is not optimized.

Both sampling-based and reparameterization-based methods provide a great boost to the search efficiency. However, the topological structure within searchable paths is neglected. This leads to the instability that lies in the topology of the reparameterized optimization problem. In addition, reparameterization-based NAS requires that the search space has to be linear, while most of the common search spaces contain nonlinear functions, limiting the range of its applications. In this work, we lift the restriction of linearity of the search space through topological simplification and propose a novel kernel normalization technique that guarantees search stability.

\section{Methods} \label{s_methods} 

In this section, we first investigate the non-linearity in DARTS' search space that prevents efficient reparameterization. After that, we introduce topological simplification as a practical solution to the issue. Finally, we analyze the instability introduced by the reparameterization and propose kernel normalization to ensure the search stability.

\subsection{The Non-linearity in DARTS Search Space}

We take the search space used in DARTS~\cite{liu2018darts} for image classification on CIFAR-10 as an example (hereafter referred to as \textit{DARTS search space}) to discuss how the non-linearity in different candidate operations causes difficulties for optimization via reparameterization.

A search cell in DARTS is a directed acyclic graph consisting of nodes and edges. A node represents a feature map while an edge represents a searchable path. The candidate operations of an edge are parameter-less $zero$, identity, average pooling, max pooling operations, and 4 convolution operations, including $3\times3$ and $5\times5$ separable convolution, as well as $3\times3$ and $5\times5$ dilated convolution. Each operation can be expressed as a combination of multiple basic modules connected in series, as shown in Table \ref{t_darts_space}.  

\begin{table*}[htb]
\centering
\caption{Candidate operations in DARTS search space and their compositions. PtwConv stands for point-wise convolution, PtwDilConv stands for point-wise dilated convolution and FC stands for $1\times 1$ convolution.}
\begin{tabular}{ c l } 
 \toprule
 Operation & Basic Modules \\
 \hline
 Zero & None \\ 
 Identity & (Identity) \\ 
 Average Pooling & (AvgPool) \\ 
 Max Pooling & (MaxPool) \\ 
 Dilated Convolution & (ReLU)-(PtwDilConv)-(FC)-(BN)  \\ 
 Separable Convolution & (ReLU)-(PtwConv)-(FC)-(BN)-(ReLU)-(PtwConv)-(FC)-(BN) \\ 
 \bottomrule
\end{tabular}
\label{t_darts_space}
\end{table*}

The reparameterization of convolutional kernels relies on the linearity of the candidate operations, while only 3 of 8 operations (i.e.\ zero, identity, and average pooling) can be represented directly by an equivalent convolutional operation. Despite being linear itself, the dilated convolutions are wrapped with ReLU and BN layers, which breaks the linearity of the entire operation. Moreover, the separable convolution operations consist of two groups of ReLU-Conv-BN combinations following the convention by~\citet{nasnet}, which further increases the non-linearity.

\subsection{Topological Simplification} 

In the previous subsection, we showed that reparameterization does not work well with DARTS search space. In contrary, the search space proposed by Single-path NAS~\cite{stamoulis2020single} can be perfectly reparameterized by sharing the non-linear modules around the convolution and only search for the linear convolution itself. This does not apply to the DARTS search space since not all candidate operations share the same form of non-linearity. We suggest that the heart of the problem is the unaligned levels of non-linearity that is blocking the module sharing and reparameterization. Therefore, we propose two basic transformations of the edge topology, namely partial module sharing and floating module sharing. By applying the transformations iteratively to simplify the topological structure of a searchable edge, we can fully leverage the power of reparameterization and accelerate the search process.

\subsubsection{Partial Module Sharing}
\label{ss_pms}

\begin{figure*}[bt]
\centering
\includegraphics[width=0.9\linewidth]{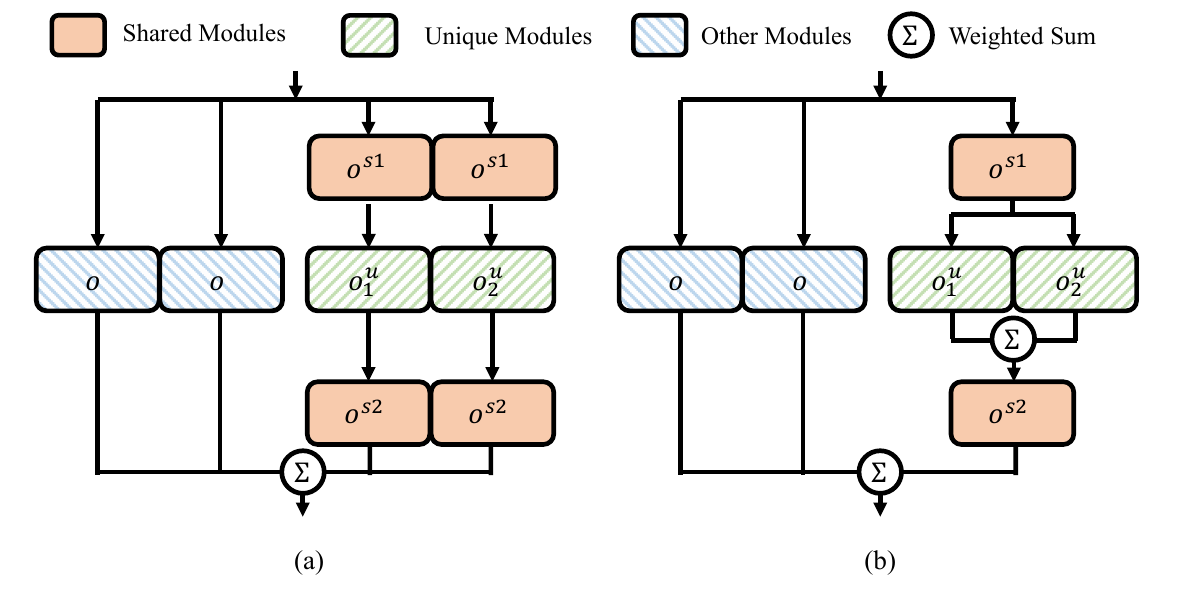}

\caption{(a) Parallel topology of an edge. (b) Hierarchical topology after merging shared modules.}
\label{fig_partial}
\end{figure*}

We propose that a searchable edge can be simplified by merging shared modules, even if the shared module only appears in a subset of candidate operations. Given $M$ candidate operations $\{o_1(x), o_2(x), \cdots, o_M(x)\}$, the vanilla operation selection process of DARTS can be expressed as a weighted sum:

\begin{equation} \label{e_vanilla_darts}
f(x)=\sum_{i=1}^{M} \beta_i o_i(x)
\end{equation}

\noindent where $\beta_i = \exp{\alpha_i} / (\sum_{j=1}^M \exp \alpha_j) $ is computed by applying softmax to the architecture parameter, representing the probability of each candidate operation. Assuming that some of the candidate operations contain shared modules, the edge could be expressed as nested functions of shared parts $o^{s1}(x), o^{s2}(x)$ and unique parts $o^{u}_i(x)$, as shown in Fig. \ref{fig_partial}(a),

\begin{equation} \label{e_shared_and_unique}
f(x)=\sum_{i\in\mathcal{I_N}}^{} \beta_i o_i(x) + \sum_{i\in\mathcal{I_S}} \beta_i o^{s2}(o^{u}_i(o^{s1}(x)))
\end{equation}

\noindent where $\mathcal{I_S}$ is thse index set of operations with shared modules and $\mathcal{I_N}$ is the index set of other operations. To isolate the shared parts, we treat the operations with shared modules as a whole operation, denoted as $o^u(x)$. Then we decompose the module into two cascaded operation selection branches, one for selecting operations in $\left\{o_i(x) | i\in\mathcal{I_N}\right\} \cup \left\{o^u(x)\right\} $ and one for selecting different operations inside $\left\{o_i(x) | i\in\mathcal{I_S}\right\}$. The modification is visualized by Fig. \ref{fig_partial}(b), and can be expressed as

\begin{gather} \label{e_two_stage_selection}
o^u(x) = \sum_{i\in\mathcal{I_S}} \beta_i^{(1)} o^u_i(x) \\
g(x) =\sum_{i\in\mathcal{I_N}} \beta_i^{(2)} o_i(x) + \beta_{\mathcal{S}}^{(2)}o^{s2}(o^{u}(o^{s1}(x)))\label{e_two_stage_selection2}
\end{gather}

\noindent where $o^u(x)$ is the expectation of the selected unique operation inside $\mathcal{I_S}$.  $\beta_i^{(1)}$ and $\beta_i^{(2)}$ are the weights of two cascaded weighted sums. $\beta_{\mathcal{S}}^{(2)}$ stands for the probability of shared operation being selected. However, learning two sets of architecture parameters for two sets of weights results in an imbalanced architecture parameter distribution since $\beta^{(1)}$ is only valid when $\beta^{(2)}_{\mathcal{S}}$ becomes dominant. Instead, we derive their values from the original weights $\beta_i$ through their probabilistic forms. The original weight stands for the probability of an operation $o_i$ being selected, 

\begin{equation}
\beta_i = p(o_i) 
\end{equation}

\noindent while the first set of new weights stands for the probability selecting one operation among all operations with shared modules, which corresponds to the conditional probability,

\begin{equation}
\beta_i^{(1)} = p(o_i|i\in\mathcal{I_S}) = \frac{p(o_i,i\in\mathcal{I_S})}{p(i\in\mathcal{I_S})} \quad \text{for} \quad  i \in \mathcal{I_S}
\end{equation}

The second set of new weights can be expressed as

\begin{equation}
\left \{
\begin{matrix}
 \beta_i^{(2)} = p(o_i) & \text{for} \quad  i \in \mathcal{I_N} \\
 \beta_{\mathcal{S}}^{(2)} = p(i\in\mathcal{I_S}) & 
\end{matrix} \right.
\end{equation}

Substituting $\beta_i^{(1)}$ and $\beta_i^{(2)}$ in Eq. \eqref{e_two_stage_selection} and \eqref{e_two_stage_selection2} with $\beta_i$, we get

\begin{gather} \label{e_partial_res}
o^u(x) = \frac{1}{p(i\in\mathcal{I_S})}\sum_{i\in\mathcal{I_S}} \beta_i o^u_i(x) \\
g(x) =\sum_{i\in\mathcal{I_N}} \beta_i o_i(x) + p(i\in\mathcal{I_S})o^{s2}(o^{u}(o^{s1}(x)))\label{e_partial_res2}
\end{gather}

\noindent where $p(i\in\mathcal{I_S})=\sum_{j\in\mathcal{I_S}}\beta_j $. According to Eq. \eqref{e_partial_res2}, the computation of $o^{s2}(x)$ is simplified as only one operation is required.

\subsubsection{Floating Module Sharing}
\label{ss_fms}

\begin{figure}[bt]
\centering
\includegraphics[width=3.5in]{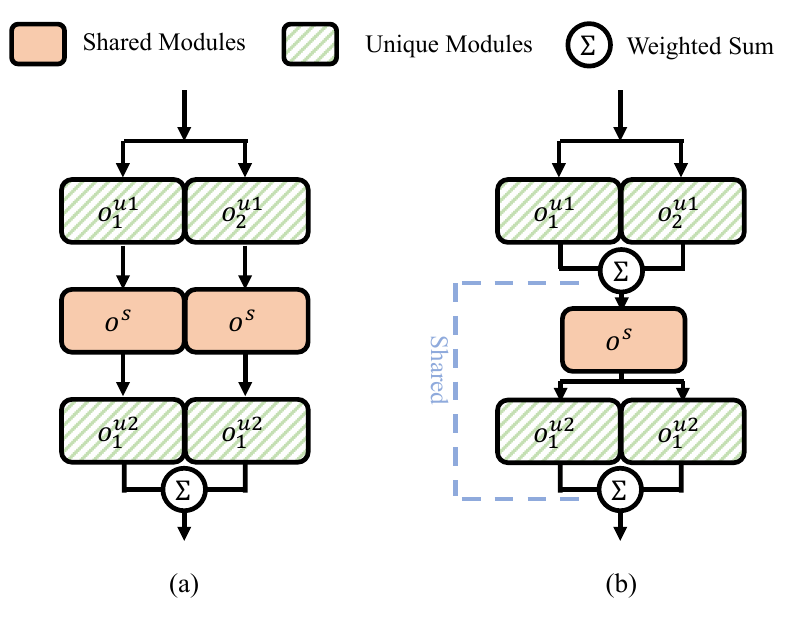}

\caption{(a) An edge with shared modules between unique modules. (b) The topology after merging shared modules. }
\label{fig_intermediate}
\end{figure}

The partial module sharing only handles the merge of sibling modules ($o^{s1}$) and modules with common child ($o^{s2}$) in a directed acyclic graph. In DARTS search space, we find that the double-stacked separable convolution operations have similar structure in between their unique parts, as illustrated in Fig. \ref{fig_intermediate}(a). In this case, we call $o^{s}$ a floating shared module. For better efficiency, we propose to rewrite the topology to two series-connected operation selection branches, with shared architecture parameters. The proposed modification is demonstrated in Fig. \ref{fig_intermediate}(b).

Specifically, the weighted sum of the original structure is

\begin{equation} \label{e_shared_and_unique_floating}
f(x)=\sum_{i} \beta_i o^{u2}_i(o^{s}(o^{u1}_i(x)))
\end{equation}

\noindent where other modules connected in parallel are omitted since the merge of floating modules does not affect them. The proposed modification can be expressed as

\begin{gather} \label{e_two_stage_series}
h(x) = \sum_{i} \beta_i o^{u1}_i(x) \\
g(x) =\sum_{i} \beta_i o^{u2}_i(o^{s}(h(x)))
\label{e_two_stage_series2}
\end{gather}

\noindent where $h(x)$ is the intermediate result of the first operation selection. As in Eq. \eqref{e_two_stage_series2}, the simplified procedure requires the function $o^{s}(h(x))$ to be evaluated only once. Another benefit is that if $o^{u1}_i$ or $o^{u2}_i$ is linear, we can merge them by reparameterization, which is not feasible if the floating modules are not shared.

\subsection{General Recursive Simplification}

\begin{figure*}[htb]
\centering
\includegraphics[width=\textwidth]{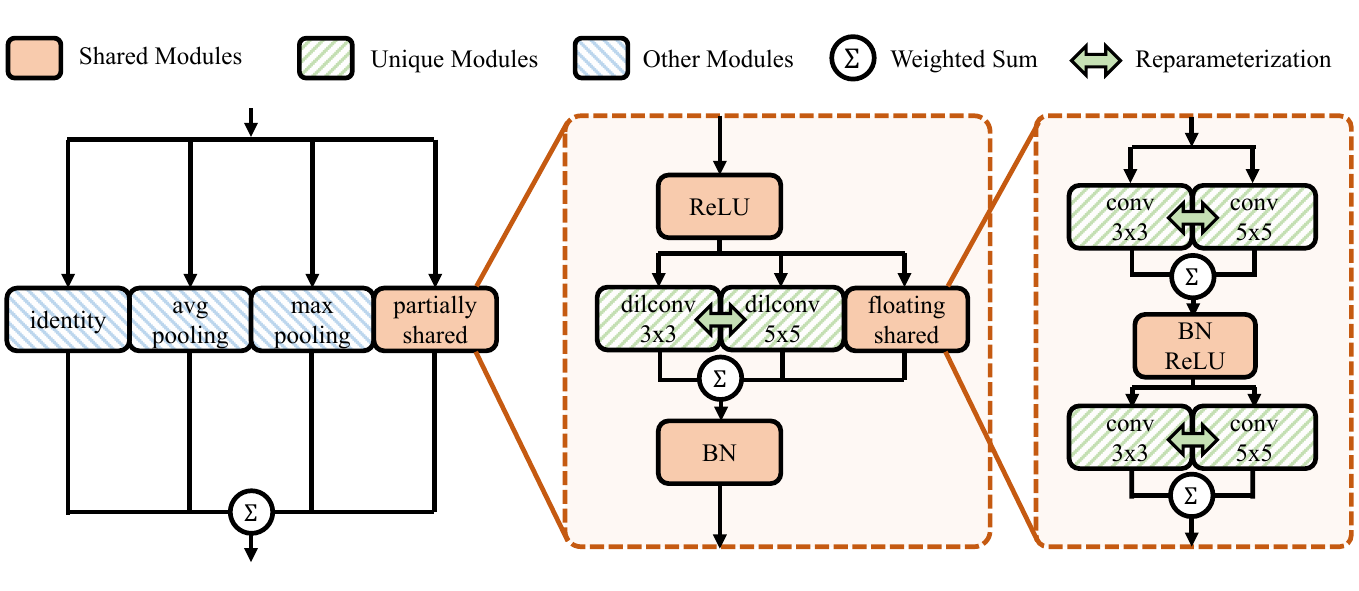}

\caption{The topology of a simplified edge in DARTS search space. }
\label{fig_dartssimp}
\end{figure*}

With two basic transformations introduced above, we can now shrink the topological structure of the search space for better efficiency. The DARTS search space can be taken as a comprehensive example to illustrate this process. As shown in Fig. \ref{fig_dartssimp}, for DARTS search space, we first group all convolution operations through partial module sharing since they share the same structure of a leading ReLU and a trailing BN module. On top of that, we apply floating module sharing to the two double-stacked separable convolution operations, resulting in shared BN and ReLU modules between two layers of convolutions. 

After simplification, the remaining unique modules for each convolution operation become adjacent to weighted sum calculations, allowing us to use reparameterization techniques to merge them into one single convolution. The number of basic modules after each simplification step is counted in Table \ref{t_module_count}. The number of convolutional and non-convolutional modules is reduced and the total number of modules needed is reduced by more than 50\% for this specific search space.

\begin{table}[htb]
\centering
\caption{The number of computations of basic modules in a DARTS edge before and after the simplification.}
\begin{tabular}{lccc}
\toprule
                        & conv & non-conv & total \\
\hline
Original search space   & 6    & 14       & 20    \\
+ Partial module sharing  & 6    & 8        & 14    \\
+ Floating module sharing & 6    & 6        & 12    \\
+ Reparameterization      & 3    & 6        & 9     \\
\bottomrule
\end{tabular}
\label{t_module_count}
\end{table}

More generally, for any given search space, topology simplification is done in a recursive manner. After applying partial module sharing to a graph $\mathcal{G}$, a subgraph representing the weighted sum of unique modules is generated, as $o^u(x)$ in Eq. \eqref{e_partial_res}. Another partial module sharing or floating module sharing transform can then be applied to the subgraph recursively until there are no shared modules to be merged.

A more precise description of this procedure is presented in Alg. \ref{alg_main}, where PMS stands for Partial Module Sharing and FMS stands for Floating Module Sharing. Reparameterization is applied to the graph after all possible PMS and FMS are done to maximize the computing efficiency.

\begin{algorithm}
\caption{Recursive Simplification}\label{alg_main}
\begin{algorithmic}[1]

\While{True}
\If{$\mathcal{G}$ has PMS structure}
    \State $\mathcal{G} \gets \text{PMS}(\mathcal{G})$
\ElsIf{$\mathcal{G}$ has FMS structure}
    \State $\mathcal{G} \gets \text{FMS}(\mathcal{G})$
\Else
    \State $\mathcal{G} \gets \text{Reparameterize}(\mathcal{G})$
    \State \textbf{break}
\EndIf
\EndWhile
\end{algorithmic}
\end{algorithm}

\subsection{Reparameterization}

After the shared modules are simplified recursively, we now focus on the remaining unique functions in Eq.\eqref{e_partial_res}, Eq.\eqref{e_two_stage_series} and Eq.\eqref{e_two_stage_series2}. In practice, the remaining unique functions in many commonly used search spaces (DARTS, NASBench201, SPOS, etc.) are either convolutional or can be expressed as an equivalent convolutional function, making them suitable for reparameterization.

MergeNAS~\cite{wang2021mergenas} shows that dilated convolution or convolutional kernels of different sizes can be rewritten as standard convolutional kernels of the same size. Following this practice, we unify different types of convolutions and denote $o^u_i(x)$ with reparameterized kernels $o^u_i(x) = K_i * x$.

Subsequently, Eq.\eqref{e_partial_res} is reparameterized as

\begin{equation}
    o^u(x) = \frac{1}{p(i\in\mathcal{I_S})}\left(\sum_{i\in\mathcal{I_S}} \beta_i K_i \right) * x
\end{equation}

Similarly, denoting $o^u_{i1}(x) = K^{(1)}_i * x$, $o^u_{i2}(x) = K^{(2)}_i * x$ , Eq.\eqref{e_two_stage_series} and Eq.\eqref{e_two_stage_series2} are reparameterized as

\begin{gather} \label{e_two_stage_series_rep}
h(x) = \left( \sum_{i} \beta_i K_i^{(1)} \right) * x \\
g(x) = \left( \sum_{i} \beta_i K_i^{(2)} \right) * o^{s}(h(x))
\label{e_two_stage_series2_rep}
\end{gather}

\subsection{Kernel Normalization}

In this section, we present an analysis to show that reparameterization-based NAS methods suffer from an increase in the instability of the search process. As our topological simplifications enable more modules to be reparameterized, the search instability is also magnified in our supernet. To solve this problem, we analyze the root cause and propose kernel normalization to guarantee a stable search of the supernet.

\begin{figure*}[htb]
	\centering
		\includegraphics[width=0.99\linewidth]{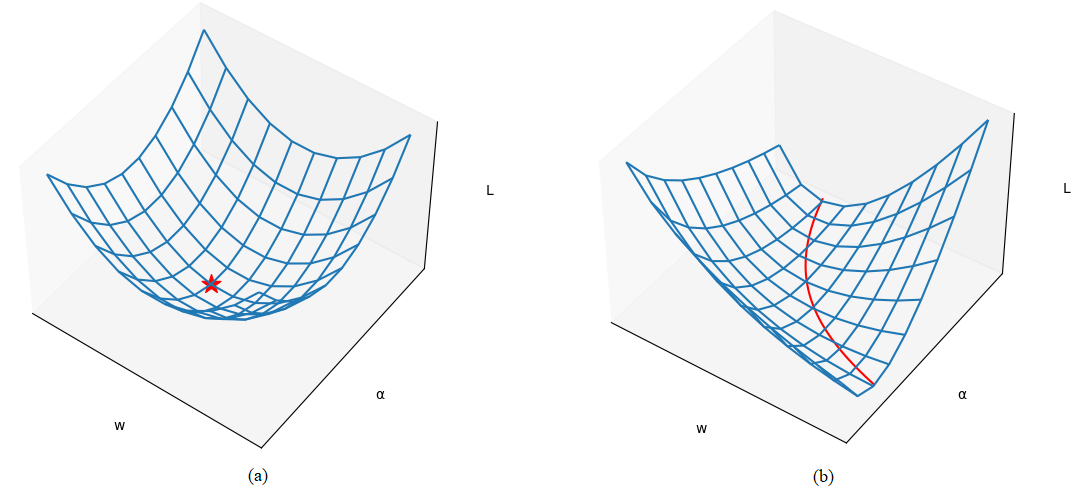}
		\caption{An illustration of the non-unique problem. (a) The loss landscape with a unique optimal $w$ and $\alpha$. (b) The loss landscape with non-unique optimal solutions. For any arbitrary $\alpha$, there exists a corresponding $w$ that makes $(\alpha, w)$ an optimal solution.}
		\label{fig_coupled}
\end{figure*}

Consider the optimization problem in Eq.\eqref{e_darts}. In order to address the issue more concretely, we introduce the following assumptions:

\begin{enumerate}
    \item The training set and the validation set are independently drawn from the same distribution.
    \item The same form of loss function is employed for $\mathcal{L}_{train}$ and $\mathcal{L}_{val}$.
    \item There exists at least one optimal solution to Eq.\eqref{e_darts}.
\end{enumerate}

Without loss of generality, we consider a local structure of the supernet which is composed of only one searchable operation, with a search space of $N$ types of activation-convolution-batch normalization blocks. Both weights and architecture parameters of other parts of the supernet are considered frozen.

\begin{equation}
    f(x; K, \alpha) = \sum_{i=1}^{N} \beta_i(\alpha) \text{BN} \left( K_i  * \text{ReLU}(x) \right)
\end{equation}

\noindent where $K = \left\{K_1, K_2, \cdots, K_N \right\}$, $\alpha = \left\{ \alpha_1, \alpha_2, \cdot, \alpha_N \right\}$, $\beta_i(\alpha) = \exp{\alpha_i} / (\sum_{j=1}^M \exp \alpha_j) $. Regardless of the specific form of convolution operations, we apply partial module sharing and reparameterization to simplify the supernet.

\begin{equation} \label{e_g_x_k_a}
    g(x; K, \alpha) = \text{BN} \left( \left( \sum_{i=1}^{N} \beta_i(\alpha) K_i \right) * \text{ReLU}(x) \right)
\end{equation}

We propose that for the reparameterized network, Any arbitrary $\alpha$ is optimal. The proof is given as follows.

Let the pair $(\alpha^*, K^*)$ be an optimal solution to Eq.\eqref{e_darts}. For any arbitrary $\alpha$, we construct $K(\alpha)=\{K_1, K_2, \cdots, K_N\}$ as

\begin{equation} \label{e_construct_k}
    K_i = K_i^* \cdot \exp (\alpha_i^* - \alpha_i) \cdot \frac{\sum_j^N \exp \alpha_j}{\sum_j^N \exp \alpha_j^*}
\end{equation}

The constructed $(\alpha, K)$ is also an optimal solution to Eq.\eqref{e_darts}. To validate that, we first show that under the construction of Eq.\eqref{e_construct_k}, Eq.\eqref{e_g_x_k_a} and the loss function are invariant for such $(\alpha, K)$.

Consider $\sum_{i=1}^{N} \beta_i(\alpha) K_i$ in Eq.\eqref{e_g_x_k_a}. Plugging in Eq.\eqref{e_construct_k}, we get

\begin{equation}
\begin{split}
     \sum_{i=1}^{N} \beta_i(\alpha) K_i & = \sum_{i=1}^{N} \frac{\exp \alpha_i}{\sum_j^N \exp \alpha_j} K_i^* \cdot \exp (\alpha_i^* - \alpha_i) \frac{\sum_j^N \exp \alpha_j}{\sum_j^N \exp \alpha_j^*} \\
     & = \sum_{i=1}^{N} \beta_i(\alpha^*) K_i^*
\end{split}
\end{equation}

\begin{equation} \label{e_g_x_k_a_invariant}
\begin{split}
     g(x; K, \alpha) & = \text{BN} \left( \left( \sum_{i=1}^{N} \beta_i(\alpha) K_i \right) * \text{ReLU}(x) \right) \\
     &= \text{BN} \left( \left(  \sum_{i=1}^{N} \beta_i(\alpha^*) K_i^* \right) * \text{ReLU}(x) \right) \\
     &= g(x; K^*, \alpha^*)
\end{split}
\end{equation}

\noindent which indicates that the output of the local structure is unchanged. As the other part of the supernet is frozen, the output of the supernet also remains invariant. As a result, the loss functions are also invariant for such $(\alpha, K)$.

\begin{gather} \label{e_l_train_1}
    {L}_{train} ( K, \alpha ) = {L}_{train} ( K^*, \alpha^* ) \\
    \label{e_l_val}
    {L}_{val} ( K, \alpha ) = {L}_{val} ( K^*, \alpha^* )
\end{gather}

After that, we prove that $K(\alpha)$ is optimal to the problem $\text{argmin}_{K}  \mathcal{L}_{train} ( K, \alpha )$ by contradiction.

Suppose, for the sake of contradiction, there exists $\hat{K}$ such that 

\begin{equation} \label{e_l_train_ue}
\mathcal{L}_{train} ( \hat{K}, \alpha ) < \mathcal{L}_{train} ( K, \alpha )\
\end{equation}

We construct $\bar{K}$ as follows

\begin{equation}
    \bar{K}_i = \hat{K}_i \cdot \exp (\alpha_i - \alpha_i^*) \frac{\sum_j^N \exp \alpha_j^*}{\sum_j^N \exp \alpha_j}
\end{equation}

Similarly, we have invariant loss functions for $(\alpha^*, \bar{K})$ by construction.

\begin{equation} \label{e_l_train_2}
    {L}_{train} ( \bar{K}, \alpha^* ) = {L}_{train} ( \hat{K}, \alpha )
\end{equation}

Combining Eq.\eqref{e_l_train_1}, \eqref{e_l_train_ue}, \eqref{e_l_train_2}, we have

\begin{equation}
\begin{split}
    {L}_{train} ( \bar{K}, \alpha^* ) & = {L}_{train} ( \hat{K}, \alpha ) \\
    & < {L}_{train} ( K, \alpha ) = {L}_{train} ( K^*, \alpha^* )
\end{split}
\end{equation}

\noindent which contradicts with the optimality of $(\alpha^*, K^*)$. Therefore, the statement $K(\alpha)=\text{argmin}_{K}  \mathcal{L}_{train} ( K, \alpha )$ is true, which means $(\alpha, K)$ satisfies the constraint in Eq.\eqref{e_darts}. And because the objective function equals the optimal value ${L}_{train} ( K, \alpha ) = {L}_{train} ( K^*, \alpha^* )$, the proposition that $(\alpha, K)$ is also an optimal solution to Eq.\eqref{e_darts} is proved.

An illustration of the loss landscape is presented in Fig.\ref{fig_coupled} to better explain the concept of bi-linearity between $w$ and $\alpha$. It is worth noting that in practice the $\alpha$ and $w$ have high numbers of dimensions. For visualization, here we represent both with one-dimensional axes. As the goal of the optimization is to find the optimal $\alpha$, the ideal landscape is that there exists a unique or finite number of optimal values of $\alpha$ as shown in Fig.\ref{fig_coupled}(a). However, in reparameterization-based supernets, the loss landscape is altered so that every $\alpha$ is optimal, as shown in Fig.\ref{fig_coupled}(b). In that case, the optimization has degenerated into a problem that does not produce meaningful solutions.

To solve the problem, we propose a straightforward way to eliminate the bi-linear coupling between $\alpha$ and $w$. Let $K_i$ be the original kernels, we normalize the kernels to obtain $\bar{K}_i$.

\begin{equation}
    \bar{K}_i = \frac{K_i - \text{mean}(K_i)}{\text{std}(K_i)}
\end{equation}

The normalized kernels are insensitive to scaling and translating of the original kernels, effectively easing the bi-linear 
 coupling issue between kernels and architecture parameters.

\section{Experiments} 
\label{s_exp}

In this section, we conduct experiments on several benchmarks to analyze the performance of \strategyname{} under different conditions. The first one is NASBench201~\cite{dong2020nasbench201}, which provides detailed evaluation information on every possible architecture in a tiny search space with 4 nodes. The simplicity of NASBench201's search space has made it a widely used standard for comparing the accuracy of searching algorithms. But also because of its simplicity, there is less room for topological simplification. Therefore, we conduct additional experiments on DARTS search space, which involves separable convolutions and double-stacked convolutions. 

By experimenting in the above two search spaces, we prove that our method does not compromise searching stability while boosting efficiency. We also conduct ablation studies to verify the effect of each component of our proposed method.

\subsection{Datasets and Benchmarks}

In this paper, we conduct experiments on several datasets and benchmarks. This section provides a detailed introduction to the datasets and benchmarks used in our experiment.

1) CIFAR-10 and CIFAR-100~\cite{krizhevsky2009learningcifar} are two datasets for the image classification task. Concretely, CIFAR-10 contains small-sized images of real-world objects in 10 categories. The dataset has 60000 images, where 50000 images are for training and 10000 are for testing. CIFAR-100 is a dataset of the same form but with richer categories. It consists of 20 super-classes and 100 sub-classes, each with 500 images for training and 100 for testing. CIFAR-10 and CIFAR-100 are small in size and can be used to evaluate searched model architectures efficiently. Therefore, they are widely used as a standard benchmark to assess NAS methods.

2) ImageNet-16-120~\cite{chrabaszcz2017downsampled} is a downsampled version of the ImageNet~\cite{russakovsky2015imagenet} classification dataset. ImageNet is a large-scale image classification dataset with 1000 categories and more than one million image samples. To better assess the performance of NAS methods and reduce computation costs, the dataset is downsampled to 16 by 16 pixels and the number of categories is reduced to 120 to form the ImageNet-16-120 subset.

3) NASBench201~\cite{dong2020nasbench201} is a benchmark with a well-studied search space that provides solid evaluation metrics on every possible architecture in it. The search space is built with repeated cells consisting of a searchable directed acyclic graph. The graph in each cell contains 4 nodes and 6 searchable paths between nodes. During the search phase, a searchable path is optimized to choose one of five candidate operations, namely zero, skip, 1x1 conv, 3x3 conv, and 3x3 average pooling. Being a relatively small search space, the total number of possible cell structures in the NASBench201 search space is 15,625. NASBench201 has evaluated each possible structure on CIFAR-10, CIFAR-100, and ImageNet-16-120 datasets. Therefore, the global optimal of the search space is already known, providing researchers with a clear reference for optimizing NAS methods.

4) DARTS~\cite{liu2018darts} also utilizes a cell-based search space for architecture searching. The search space is larger than that of NASBench201 and contains more complex candidate operations. Therefore, the search space serves as a more challenging benchmark for NAS methods. As the number of possible cell structures increases rapidly with respect to the number of nodes and candidate operations, the search space is not fully evaluated and the global optimal of it is unknown. However, we can still assess the performance of NAS methods by their classification accuracy, search time, and memory usage.

\subsection{Experiments with NASBench201} 

We apply \strategyname{} as a plug-in to DARTS and its state-of-the-art variants, including PC-DARTS~\cite{xu2019pc}, $\beta$-DARTS~\cite{ye2022beta} and GDAS~\cite{dong2019gdas}. During the search phase, we optimize the supernet with an SGD optimizer for model parameters and an Adam optimizer for architecture parameters. The learning rate of model parameters is set to $0.025$ initially and decreases according to a cosine annealing schedule. The learning rate of architecture parameters is set to $3.0\times 10^{-4}$. DARTS is known to have low robustness on the NASBench201 search space, as reported by~\citet{zela2019understanding}. Therefore, we increase the weight decay to $1\times 10^{-2}$ for DARTS following~\citet{wang2021mergenas}. 

As shown in Table \ref{t_result_201}, We use pre-computed validation scores on CIFAR-10, CIFAR-100~\cite{krizhevsky2009learningcifar} and ImageNet-16-120~\cite{chrabaszcz2017downsampled} provided by the benchmark to evaluate the performance of our search strategy. For each dataset, the top-1 classification accuracy of the validation set and the test set are reported. The second row of each method displays the standard deviation of corresponding metrics from five runs.

To visualize both search efficiency and accuracy altogether, we propose a new metric named \textit{normalized cost} $C$ to measure the overall search cost of each method as follows.

\begin{equation}
    C = \frac{1}{2} ( \frac{M}{M_{min}} + \frac{T}{T_{min}} )
\end{equation}

\noindent where $M$ stands for memory consumption and $T$ stands for time cost. $M_{min}$ denotes the lowest memory consumption among all methods and $T_{min}$ denotes the minimum time cost among all methods. In general, a lower normalized cost represents better search efficiency. 

\begin{figure}[htb]
	\centering
		\includegraphics[width=0.9\linewidth]{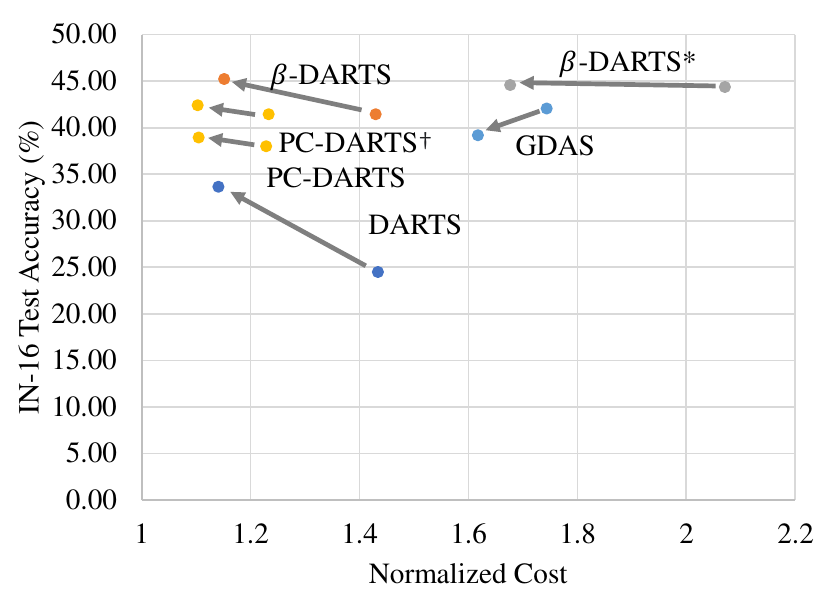}
		\caption{Comparison on NASBench201. The data points are colored by types of baseline methods. The arrows indicate the changes in search efficiency and accuracy before and after applying our strategy. A lower normalized cost indicates better efficiency and a higher test score indicates better accuracy. Therefore, points closer to the top left corner represent better overall performances. }
		\label{fig_compare_nasbench}
\end{figure}

A scatter chart illustrating the relationship between normalized cost and test set accuracy on ImageNet-16 is presented in Fig.\ref{fig_compare_nasbench}. Comparing the differences before and after plugging in our search strategy, we see improvements in both search efficiency and accuracy for most methods. Among all experiments, the best overall performance is achieved when applying \strategyname{} to $\beta$-DARTS and PC-DARTS. The topological simplification and reparameterization strategy bring at least a 7\% reduction in normalized search cost. For DARTS and $\beta$-DARTS, the cost reduction is around 20\%. We also observe a 9.16\% improvement in IN-16 test accuracy and a reduced standard deviation when applying our strategy to plain DARTS. Interestingly, not all baseline methods receive that much benefit from our plugin in terms of accuracy and stability. When combined with GDAS, the search accuracy slightly drops while the efficiency is improved. This suggests that the accuracy increase is affected by the properties of the baseline model. As a key factor affecting the accuracy of our method, the kernel normalization technique can also bring a gap between searching and retraining. More accuracy improvement is observed on the original DARTS methods while less is observed on sophisticated methods like GDAS and $\beta$-DARTS with extended epochs, which utilize techniques that improve search stability at the cost of reduced performance.

In summary, our method acts as a plugin to one-shot NAS compatible with a wide range of state-of-the-art searching strategies and enhances both efficiency and accuracy to reach a better overall performance.

\begin{table*}[htb]
\centering

\caption{Comparison on NASBench201. We report the mean and standard deviation of five runs of all comparison methods. Evaluation scores are pre-computed by NASBench201. The optimal value is reported by NASBench201 by evaluating every possible architecture. The original work of $\beta$-DARTS~\cite{ye2022beta} uses a search schedule of 100 epochs instead of the standard setting of 50 epochs in NASBench201. For a fair comparison, we conduct experiments using both settings. $\beta$-DARTS denotes searching for 50 epochs and $\beta$-DARTS* denotes searching with the settings aligned with the original work. $\dagger$ denotes combining $\beta$ regularization proposed by $\beta$-DARTS with PC-DARTS. Please refer to Fig.\ref{fig_compare_nasbench} for a more straightforward visualization.}

\begin{tabular}{@{}cccccccccc@{}}
\toprule
\multicolumn{1}{l}{} & \multirow{2}{*}{Memory/MB} & \multirow{2}{*}{Time/s} & \multicolumn{2}{c}{CIFAR10} & \multicolumn{2}{c}{CIFAR100} & \multicolumn{2}{c}{IN-16} & \multirow{2}{*}{\begin{tabular}[c]{@{}c@{}}Normalized\\      Cost\end{tabular}} \\
\multicolumn{1}{l}{} &  &  & val & test & val & test & val & test &  \\ \midrule
\multirow{2}{*}{DARTS\cite{liu2018darts}} & 3626 & 11386 & 78.20 & 82.79 & 52.25 & 52.25 & 25.10 & 24.50 & \multirow{2}{*}{1.43} \\
 & - & $\pm$137 & $\pm$5.07 & $\pm$4.93 & $\pm$7.33 & $\pm$7.44 & $\pm$5.97 & $\pm$6.15 &  \\
\multirow{2}{*}{DARTS+Ours} & 3006 & \topone{8620} & 85.26 & 89.17 & 62.56 & 62.68 & 33.74 & 33.66 & \multirow{2}{*}{1.14} \\
 & - & $\pm$157 & $\pm$2.63 & $\pm$2.26 & $\pm$4.44 & $\pm$4.80 & $\pm$4.99 & $\pm$5.25 &  \\
\multirow{2}{*}{$\beta$-DARTS\cite{ye2022beta}} & 3626 & 11318 & 90.20 & \toptwo{93.76} & 70.71 & 71.11 & 40.78 & 41.44 & \multirow{2}{*}{1.43} \\
 & - & $\pm$73 & $\pm$0.00 & $\pm$0.00 & $\pm$0.00 & $\pm$0.00 & $\pm$0.00 & $\pm$0.00 &  \\
\multirow{2}{*}{$\beta$-DARTS+Ours} & 3006 & \toptwo{8800} & \topone{91.22} & \topone{94.05} & \topone{72.68} & \topone{72.87} & \topone{45.14} & \topone{45.24} & \multirow{2}{*}{1.15} \\
 & - & $\pm$124 & $\pm$0.45 & $\pm$0.41 & $\pm$1.00 & $\pm$0.79 & $\pm$1.59 & $\pm$1.55 &  \\
\multirow{2}{*}{$\beta$-DARTS*} & 3626 & 22365 & 90.33 & 93.24 & 71.18 & \toptwo{71.27} & 44.37 & 44.38 & \multirow{2}{*}{2.07} \\
 & - & $\pm$139 & $\pm$1.76 & $\pm$1.51 & $\pm$2.44 & $\pm$2.32 & $\pm$2.33 & $\pm$2.77 &  \\
\multirow{2}{*}{$\beta$-DARTS*+Ours} & 3006 & 17854 & \toptwo{90.62} & 93.36 & 70.91 & 70.85 & \toptwo{44.50} & \toptwo{44.60} & \multirow{2}{*}{1.68} \\
 & - & $\pm$261 & $\pm$0.78 & $\pm$0.81 & $\pm$1.85 & $\pm$1.84 & $\pm$1.95 & $\pm$1.87 &  \\
\multirow{2}{*}{PC-DARTS\cite{xu2019pc}} & 2500 & 11987 & 88.15 & 91.70 & 66.78 & 66.82 & 37.32 & 38.01 & \multirow{2}{*}{1.23} \\
 & - & $\pm$100 & $\pm$1.29 & $\pm$0.95 & $\pm$1.80 & $\pm$2.15 & $\pm$1.59 & $\pm$1.88 &  \\
\multirow{2}{*}{PC-DARTS+Ours} & \topone{2344} & 10430 & 89.79 & 93.12 & 69.90 & 70.45 & 39.00 & 38.94 & \multirow{2}{*}{\topone{1.10}} \\
 & - & $\pm$159 & $\pm$0.66 & $\pm$0.38 & $\pm$0.89 & $\pm$0.98 & $\pm$3.00 & $\pm$3.08 &  \\
\multirow{2}{*}{PC-DARTS$^{\dagger}$} & 2500 & 12066 & 90.20 & \toptwo{93.76} & 70.71 & 71.11 & 40.78 & 41.44 & \multirow{2}{*}{1.23} \\
 & - & $\pm$82 & $\pm$0.00 & $\pm$0.00 & $\pm$0.00 & $\pm$0.00 & $\pm$0.00 & $\pm$0.00 &  \\
\multirow{2}{*}{PC-DARTS$^{\dagger}$+Ours} & \topone{2344} & 10392 & 90.12 & 93.62 & 70.76 & 71.08 & 41.99 & 42.40 & \multirow{2}{*}{\topone{1.10}} \\
 & - & $\pm$203 & $\pm$0.12 & $\pm$0.18 & $\pm$0.31 & $\pm$0.53 & $\pm$1.32 & $\pm$1.40 &  \\
\multirow{2}{*}{GDAS\cite{dong2019gdas}} & 2540 & 20720 & 89.86 & 93.58 & \toptwo{71.33} & 70.60 & 42.07 & 42.06 & \multirow{2}{*}{1.74} \\
 & - & $\pm$268 & $\pm$0.08 & $\pm$0.09 & $\pm$0.05 & $\pm$0.31 & $\pm$1.41 & $\pm$1.04 &  \\
\multirow{2}{*}{GDAS+Ours} & 2582 & 18389 & 88.77 & 92.08 & 67.58 & 67.33 & 39.04 & 39.20 & \multirow{2}{*}{1.62} \\
 & - & $\pm$236 & $\pm$0.16 & $\pm$0.18 & $\pm$0.03 & $\pm$0.16 & $\pm$0.28 & $\pm$0.71 &  \\ \midrule
Optimal & - & - & 91.61 & 94.37 & 73.49 & 73.51 & 46.77 & 47.31 & - \\ \bottomrule
\end{tabular}
\label{t_result_201}
\end{table*}

\subsection{Experiments with DARTS Search Space} 
\label{s_darts_exp}

We compare the performance before and after plugging our method into vanilla DARTS and $\beta$-DARTS with DARTS search space. Our implementation and experiment setup for the DARTS search space follows NNI~\cite{Microsoft_Neural_Network_Intelligence_2021}, an open-source framework for neural architecture search and parameter tuning. In the search phase, the supernet is initialized to have 8 layers and trained on the CIFAR-10 dataset for 50 epochs. In the retraining phase, the found architecture is extended to 20 layers and trained for 600 epochs. Other hyperparameters remain the same as NASBench201.

As shown in Table \ref{t_result_darts}, our final architecture has a similar or better retraining performance to the baseline model, while saving 40\% of memory and 0.1 hours of search time. The improvement in search time is not as remarkable as memory, which is caused by the overhead introduced by the simplification process. We also compute the normalized cost as introduced in the previous experiment, which indicates that \strategyname{} brings a considerable amount of efficiency boost on DARTS search space. The accuracy of DARTS after applying TopoNAS is improved by 1.2\%, while that of $\beta$-DARTS is nearly the same (-0.1\%), similar to the results on NASBench201. The reason for this is that $\beta$-DARTS itself offers robust regularization to stabilize the search process, therefore the effectiveness of layer normalization is not very pronounced when used in conjunction with it.

\begin{table*}[htb]
\centering
\caption{Comparison on DARTS search space. Time consumption and peak memory consumption are recorded during the search phase. The top-1 and top-5 accuracy on the CIFAR-10 test set are evaluated during the retraining phase. }
\begin{tabular}{@{}cccccc@{}}
\toprule
\multicolumn{1}{l}{} & \multirow{2}{*}{Time/h} & \multirow{2}{*}{Memory/MB} & \multicolumn{2}{c}{Accuracy} & \multirow{2}{*}{\begin{tabular}[c]{@{}c@{}}Normalized\\      Cost\end{tabular}} \\
\multicolumn{1}{l}{} &  &  & top1 & top5 &  \\ \midrule
DARTS & 3.4 & 19752 & 94.94 & 99.9 & 1.35 \\
DARTS+Ours & \topone{3.3} & \topone{11760} & \topone{96.17} & 99.9 & \topone{1.00} \\ \midrule
$\beta$-DARTS & 3.4 & 19752 & \topone{96.71} & 99.9 & 1.35 \\
$\beta$-DARTS+Ours & \topone{3.3} & \topone{11760} & 96.58 & 99.9 & \topone{1.00} \\ \bottomrule
\end{tabular}
\label{t_result_darts}
\end{table*}

\begin{table*}[]
\centering
\caption{ \todo{fold table} Ablation study on module sharing (+Sharing), kernel normalization (+KN) and reparameterization (+Reparam). The top1 and top2 metrics are marked with red and blue colors.}
\begin{tabular}{@{}crcccccccc@{}}
\toprule
\multirow{2}{*}{Method} & \multirow{2}{*}{Memory/MB} & \multirow{2}{*}{Time/s} & \multicolumn{2}{c}{CIFAR10} & \multicolumn{2}{c}{CIFAR100} & \multicolumn{2}{c}{IN-16} & \multirow{2}{*}{\begin{tabular}[c]{@{}c@{}}Normalized\\      Cost\end{tabular}} \\
 &  &  & val & test & val & test & val & test &  \\ \midrule
\multirow{15}{*}{DARTS}
  & \multicolumn{1}{l}{+None} \\
 & 3626 & 11386 & 78.20 & 82.79 & 52.25 & 52.25 & 25.10 & 24.50 & 1.43 \\
 & - & $\pm$137 & $\pm$5.07 & $\pm$4.93 & $\pm$7.33 & $\pm$7.44 & $\pm$5.97 & $\pm$6.15 &  \\
  & \multicolumn{1}{l}{+Sharing} \\
 & 3414 & 10761 & 50.60 & 61.01 & 24.58 & 24.69 & 17.27 & 16.57 & 1.35 \\
 & - & $\pm$130 & $\pm$19.86 & $\pm$15.43 & $\pm$17.80 & $\pm$17.63 & $\pm$8.45 & $\pm$8.04 &  \\
  & \multicolumn{1}{l}{+Sharing+KN} \\
 & 3422 & 11774 & \toptwo{83.44} & \toptwo{88.11} & \toptwo{60.11} & \toptwo{59.77} & \toptwo{32.51} & \toptwo{32.39} & 1.41 \\
 & - & $\pm$566 & $\pm$5.13 & $\pm$3.77 & $\pm$7.27 & $\pm$7.32 & $\pm$8.02 & $\pm$8.80 &  \\
  & \multicolumn{1}{l}{+Sharing+Reparam} \\
 & \topone{3000} & \topone{8145} & 43.83 & 58.95 & 18.83 & 19.33 & 16.52 & 15.97 & \topone{1.11} \\
 & - & $\pm$265 & $\pm$15.81 & $\pm$11.05 & $\pm$14.61 & $\pm$14.51 & $\pm$5.80 & $\pm$5.38 &  \\
  & \multicolumn{1}{l}{+All} \\
 & \toptwo{3006} & \toptwo{8620} & \topone{85.26} & \topone{89.17} & \topone{62.56} & \topone{62.68} & \topone{33.74} & \topone{33.66} & \toptwo{1.14} \\
 & - & $\pm$157 & $\pm$2.63 & $\pm$2.26 & $\pm$4.44 & $\pm$4.80 & $\pm$4.99 & $\pm$5.25 &  \\ \midrule
\multirow{15}{*}{$\beta$-DARTS} 
  & \multicolumn{1}{l}{+None} \\
 & 3626 & 11318 & 90.20 & 93.76 & 70.71 & 71.11 & 40.78 & 41.44 & 1.43 \\
 & - & $\pm$73 & $\pm$0.00 & $\pm$0.00 & $\pm$0.00 & $\pm$0.00 & $\pm$0.00 & $\pm$0.00 &  \\
  & \multicolumn{1}{l}{+Sharing} \\
 & 3414 & 11039 & 88.99 & 92.06 & 68.02 & 68.15 & 40.95 & 41.07 & 1.37 \\
 & - & $\pm$181 & $\pm$3.26 & $\pm$2.57 & $\pm$4.63 & $\pm$4.53 & $\pm$6.42 & $\pm$6.06 &  \\
  & \multicolumn{1}{l}{+Sharing+KN} \\
 & 3422 & 11251 & \topone{91.34} & \toptwo{94.01} & \toptwo{72.28} & \toptwo{72.46} & \topone{45.99} & \topone{45.35} & 1.38 \\
 & - & $\pm$431 & $\pm$0.16 & $\pm$0.23 & $\pm$0.85 & $\pm$0.72 & $\pm$0.29 & $\pm$0.86 &  \\
  & \multicolumn{1}{l}{+Sharing+Reparam} \\
 & \topone{3000} & \topone{8449} & 86.25 & 89.22 & 62.29 & 62.08 & 36.08 & 35.61 & \topone{1.13} \\
 & - & $\pm$328 & $\pm$2.65 & $\pm$2.51 & $\pm$5.45 & $\pm$5.56 & $\pm$5.17 & $\pm$5.47 &  \\
  & \multicolumn{1}{l}{+All} \\
 & \toptwo{3006} & \toptwo{8800} & \toptwo{91.22} & \topone{94.05} & \topone{72.68} & \topone{72.87} & \toptwo{45.14} & \toptwo{45.24} & \toptwo{1.15} \\
 & - & $\pm$124 & $\pm$0.45 & $\pm$0.41 & $\pm$1.00 & $\pm$0.79 & $\pm$1.59 & $\pm$1.55 &  \\ \bottomrule
\end{tabular}

\label{t_ablation_more}
\end{table*}

\subsection{Ablation on Module Contribution}

In Table \ref{t_ablation_more}, we experiment with DARTS and $\beta$-DARTS with different configurations on NASBench201, to verify the effect of individual parts of our methods. The search cost and retraining performance on NASBench201 are listed in the table, as well as the normalized cost. 

The same trend is observed for both DARTS and $\beta$-DARTS. The module-sharing strategy and reparameterization both contribute to cost reduction. The module sharing brings a 6\% reduction to the normalized cost and the subsequent reparameterization, which is made possible only by the sharing strategy, brings 18\% more reduction to the normalized cost. However, without kernel normalization, both baseline strategies show a decrease in test accuracy. IN-16 test accuracy drops by 35\% for DARTS and 14\% for $\beta$-DARTS. After applying kernel normalization, the search cost slightly increases while the search accuracy is greatly stabilized. The kernel normalization technique adds only a small amount of computation at the level of the convolution kernel, thus has a small impact on normalized cost (+0.03). With kernel normalization, both baseline methods gain accuracy (+17.69 for DARTS and +9.63 for $\beta$-DARTS) and outperform the original method. The configuration with all plugin modules enabled (+All) achieves the best overall performance and is better compared to the baseline (+None) in terms of both accuracy and search cost.

In summary, the experiments demonstrate that the module-sharing strategy is effective in improving search efficiency while the kernel normalization technique works well for stabilizing the search result.

\subsection{Ablation on Search Efficiency}

\begin{table}[htb]
\centering
\caption{Ablations on different simplification methods applied on DARTS baseline, where PMS stands for Partial Module Sharing (Sec.~\ref{ss_pms}) and FMS stands for Floating Module Sharing (Sec.~\ref{ss_fms}).}
\begin{tabular}{@{}lllllll@{}}
\toprule
 & \multicolumn{2}{c}{Memory(MiB)} & \multicolumn{2}{c}{Time(h)} & \multicolumn{2}{c}{Normalized Cost} \\ \midrule
DARTS & 23498 &  & 24.42 &  & 2.37 &  \\
+   PMS & 15840 & -33\% & 23.63 & -3\% & 1.84 & -22\% \\
+   FMS & 12302 & -48\% & 14.71 & -40\% & 1.31 & -45\% \\
\textbf{+   Both} & \textbf{7672} & \textbf{-67\%} & \textbf{14.59} & \textbf{-40\%} & \textbf{1.00} & \textbf{-58\%} \\ \bottomrule
\end{tabular}
\label{t_ablation}
\end{table}

\begin{figure*}[htb]
	\centering
		\includegraphics[width=0.75\linewidth]{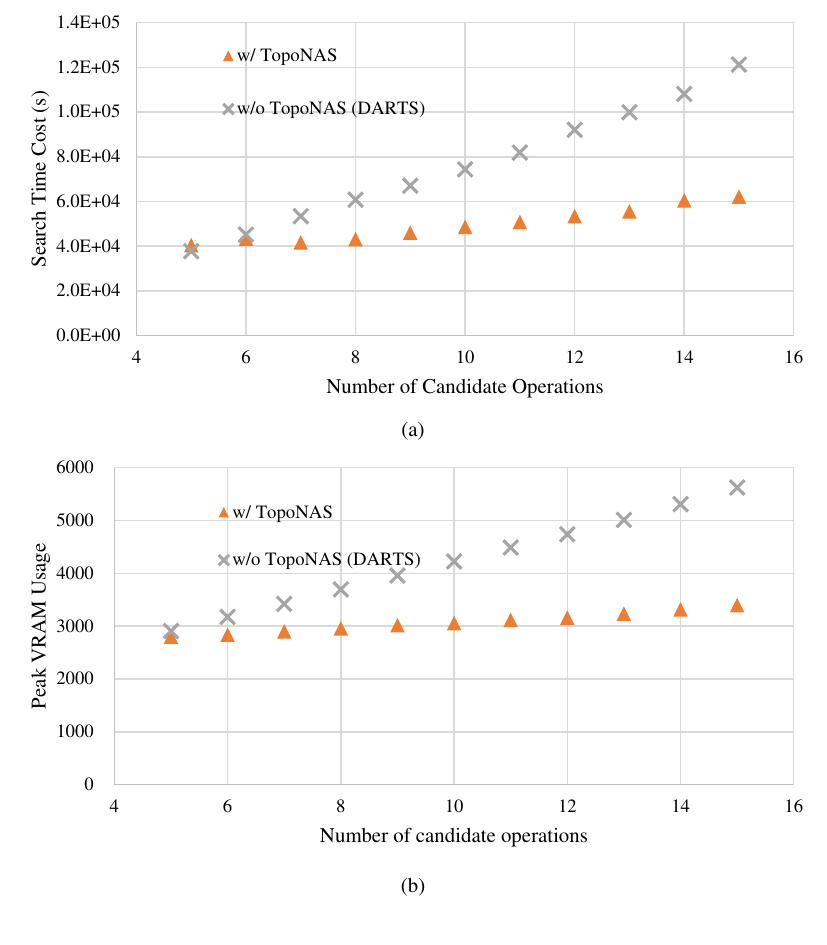}
		\caption{ Ablation on search efficiency under different numbers of candidate operations. The margin between TopoNAS and the baseline method enlarges as the number of candidate operations increases. }
		\label{fig_large_space}
\end{figure*}

To identify the difference between the effect of partial module sharing (PMS) and floating module sharing (FMS) transformations on search efficiency, we isolate the partial module sharing and floating module sharing transformations and conduct ablation studies as shown in Table \ref{t_ablation}. The primary distinction between PMS and FMS is found in their handling of complex candidate operations, such as double-stacked convolutions. Therefore, the ablation study is conducted on an exaggerated search space with 5 separable dilated convolutions and 7 double-stacked separable convolutions, to better showcase the difference. As the search space becomes larger, it is impractical to evaluate the result of each run completely. Therefore, we evaluate elapsed time searching for a single epoch and estimate the full search time by extrapolation. The search is conducted on a single NVIDIA GeForce RTX 3090 GPU. 

PMS alone provides a 33\% reduction in memory usage but has no significant impact on search time (-3\%). The double-stacked separable convolutions in the search space are the key factor to time cost and can not be fully simplified solely by PMS. FMS works well simplifying double-stacked convolutions and provides 48\% memory reduction and 40\% search time reduction by itself. However, FMS has no effect on other convolution modules that do not contain floating structures. Working together, PMS and FMS can fully simplify the topological structure and reduce the normalized cost by 58\%, substantially saving both memory and search time.

An ablation study is also conducted under different numbers of candidate operations to assess how efficiency changes relative to the size of the search space. The basic search space consists of three parameterless operations, namely skip, max pooling, and average pooling. Various separable dilated convolutions and double-stacked separable convolutions are added to create search spaces ranging from 4 to 16 candidate operations. As shown in Fig. \ref{fig_large_space}, as we add more convolution operations of different forms and sizes, the search time and VRAM usage of the baseline method grows rapidly. After applying TopoNAS to the baseline, the growth of search costs becomes more gradual. In search spaces with small numbers of candidate operations, the improvement of search efficiency is small as the overhead brought by the simplification process may cancel out the efficiency improvement it brings. However, the slope of the growth indicates that TopoNAS brings more benefits to search efficiency for larger search spaces.

\subsection{Limitations}

From the theoretical analysis and experiment results, \strategyname{} brings a valid boost to the search efficiency of gradient-based NAS. However, we also acknowledge several limitations of our method. 1) The kernel normalization proves to be a simple yet effective solution to the parameter bi-linear coupling issue. However, it may bring inconsistency between searching and retraining. 2) TopoNAS may bring extra overhead on simple search spaces, leading to less effectiveness in improving NAS efficiency. We believe these limitations can be mitigated by exploring more robust normalization or regularization manners.

\section{Conclusion} \label{s_conclusion} 

In this paper, we present \strategyname{}, a novel search strategy for one-shot NAS that minimizes the memory requirement and improves the efficiency during the search process. \strategyname{} simplifies the structure of the search space with module sharing in a recursive manner, reducing the computation needed to perform an architecture search. We also stabilize the search process by identifying the degradation issue of the reparameterized network and proposing the kernel normalization technique. In summary, \strategyname{} has the advantage of strong compatibility and stability. Extensive experiments on benchmark datasets also validate the accuracy and superior memory and time efficiency of our method. However, we find that the limitations of the proposed strategy mainly lie in the overhead of small search spaces and the gap between searching and retraining. Our future work will be focused on optimizing the overhead and finding more elegant solutions to the parameter bi-linear coupling issue.

\bibliographystyle{IEEEtran}
\bibliography{refs}

\vfill

\end{document}